\documentclass{article}

\usepackage{arxiv}

\usepackage[utf8]{inputenc} 
\usepackage[T1]{fontenc}    
\usepackage{hyperref}       
\usepackage{url}            
\usepackage{booktabs}       
\usepackage{amsfonts}       
\usepackage{nicefrac}       
\usepackage{lipsum}		
\usepackage{graphicx}
\usepackage{natbib}
\usepackage{doi}

\usepackage{listings}
\title{Modeling Arbitrarily Applicable Relational Responding with the Non-Axiomatic Reasoning System: A Machine Psychology Approach}

\date{} 					

\author{ \href{https://orcid.org/0000-0001-5547-3866}{\includegraphics[scale=0.06]{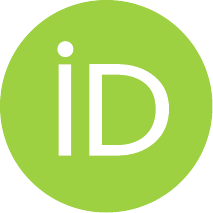}\hspace{1mm}Robert ~Johansson} \\
	Department of Psychology\\
	Stockholm University\\
	Stockholm, Sweden \\
	\texttt{robert.johansson@psychology.su.se} \\
}

\renewcommand{\headeright}{}
\renewcommand{\undertitle}{}
\renewcommand{\shorttitle}{AARR with NARS}

\hypersetup{
pdftitle={AARR with NARS},
pdfsubject={q-bio.NC, q-bio.QM},
pdfauthor={Robert ~Johansson},
pdfkeywords={First keyword, Second keyword, More},
}

\begin{document}
\maketitle

\begin{abstract}
Arbitrarily Applicable Relational Responding (AARR) is a cornerstone of human language and reasoning, referring to the learned ability to relate symbols in flexible, context-dependent ways. In this paper, we present a novel theoretical approach for modeling AARR within an artificial intelligence framework using the Non-Axiomatic Reasoning System (NARS). NARS is an adaptive reasoning system designed for learning under uncertainty. By integrating principles from Relational Frame Theory — the behavioral psychology account of AARR — with the reasoning mechanisms of NARS, we conceptually demonstrate how key properties of AARR (mutual entailment, combinatorial entailment, and transformation of stimulus functions) can emerge from NARS’s inference rules and memory structures. Two theoretical experiments illustrate this approach: one modeling stimulus equivalence and transfer of function, and another modeling complex relational networks involving opposition frames. In both cases, the system logically demonstrates the derivation of untrained relations and context-sensitive transformations of stimulus significance, mirroring established human cognitive phenomena. These results suggest that AARR — long considered uniquely human — can be conceptually captured by suitably designed AI systems, highlighting the value of integrating behavioral science insights into artificial general intelligence (AGI) research.
\end{abstract}


\section{Introduction}
Human intelligence is marked by an extraordinary capacity for symbolic reasoning — the ability to understand and manipulate symbols (words, ideas, abstract concepts) and their relationships in a flexible manner. An aspect of this flexibility is the capability to derive new relationships between symbols without direct training, purely based on their contextual relations. In cognitive and behavioral psychology, this phenomenon is captured by the concept of Arbitrarily Applicable Relational Responding (AARR), which underlies human language and higher cognition \citep{rft2001, hayes2021relating}. AARR refers to the learned behavior of relating stimuli in arbitrary ways (not dictated by the physical properties of the stimuli, but by contextual cues and social learning). For example, once a child learns that the spoken word “dog” refers to an actual furry pet, the child responds to the word as if it is functionally equivalent to the animal itself — experiencing excitement or happiness when hearing the word, similar to encountering the dog. Such symbolic equivalence is not determined by physical similarity but by relational learning. Derived relational responding of this type is considered a hallmark of human language and reasoning, enabling everything from understanding metaphors to performing complex analogies.

While humans readily perform AARR, instantiating this ability in artificial intelligence (AI) systems remains a formidable challenge. Traditional symbolic AI systems typically rely on explicitly programmed logic rules or axioms, and machine learning systems (like deep neural networks) often require vast amounts of data and struggle with extrapolating knowledge in the absence of direct examples. Achieving human-like symbolic reasoning in a machine calls for an approach that can learn relational patterns from a few examples and generalize them in a context-sensitive way, much as humans do. In other words, we seek an AI that can learn how to relate rather than being pre-programmed with all possible relations.

In this paper, we propose that AARR can be effectively modeled within a particular AI framework known as the Non-Axiomatic Reasoning System (NARS). NARS is an AI reasoning architecture designed to operate under the real-world constraints of insufficient knowledge and resources (i.e., it does not assume a closed, complete set of axioms or unlimited processing power) \citep{wang2013nalbook, wang2022intelligence}. Instead of a fixed logic, NARS uses an adaptive logic (Non-Axiomatic Logic, NAL) that allows it to learn from experience, update its beliefs probabilistically, and make plausible inferences even when knowledge is incomplete. These features make NARS a strong candidate for modeling the emergent, learned relations that characterize AARR.

The key contribution of this work is to demonstrate a computational method for describing human-like symbolic reasoning (AARR) in a machine by utilizing NARS’s capabilities. We integrate theoretical insights from Relational Frame Theory (RFT) \citep{rft2001, hayes2021relating} — the behavioral theory that explicates AARR — with the algorithmic machinery of NARS. In doing so, we show that an AI system can learn and derive relationships among symbols in a manner analogous to human relational learning. This integration provides a novel framework for studying and implementing cognitive phenomena like language and abstract reasoning in AI. Importantly, our approach goes beyond purely mechanistic or narrow AI methods: rather than training a black-box neural network on vast relational datasets, we employ a functional approach grounded in how relations are learned and used by humans \citep{johansson2024empirical}. This allows the system to capture the contextual control and generalizability of human relational responding.

This integrative approach aligns with the broader interdisciplinary perspective of \textit{Machine Psychology} \citep{johansson2024empirical, johansson2024machine}, which systematically applies principles from learning psychology — such as operant conditioning, generalized identity matching, and functional equivalence — to artificial intelligence architectures, aiming to replicate increasingly complex cognitive phenomena in machines (See Table \ref{table_empirical_studies} for an overview of how the present research fits with previously conducted studies).

We validate our approach with two experimental paradigms inspired by human studies. The first is a stimulus equivalence task involving three groups of stimuli and tests for derived symmetric and transitive relations, as well as a demonstration of the transformation of stimulus function (e.g., if one stimulus in a set is given a certain meaning or consequence, the others derived to be equivalent to it also reflect that meaning) \citep{hayes1987stimulus}. The second is an oppositional relational network task, where the system learns a network of “opposite” relations (a case of a more complex relational frame) and we examine how this leads to emergent relations and transformations of function consistent with what is observed in human experiments on relational framing of opposites \citep{roche2000contextual}.

The remainder of this article is organized as follows. In Section 2, we provide background on Arbitrarily Applicable Relational Responding and its basis in Relational Frame Theory, an overview of the Non-Axiomatic Reasoning System, as well as a section on Machine Psychology - our research approach. In Section 3, we discuss related work and contrast our approach with other efforts in AI and cognitive modeling. In Section 4, we detail how AARR can be modeled within NARS, describing the representational and algorithmic alignment between RFT concepts and NARS mechanisms. Section 5 outlines the methodology of our theoretical experiments, and Section 6 presents the results, demonstrating that NARS can indeed exhibit key properties of AARR. Finally, in Section 7 we discuss the implications of these findings for AI and cognitive science, and conclude with future directions for this interdisciplinary line of research.

\section{Background}

\subsection{Arbitrarily Applicable Relational Responding and Relational Frame Theory}
Arbitrarily Applicable Relational Responding (AARR) is a concept from behavioral psychology that refers to a general pattern of learned behavior: responding to the relation between stimuli rather than just the stimuli themselves, and doing so in a way that is not determined by the stimuli’s physical properties but by contextual cues and history of reinforcement \citep{rft2001, hayes2021relating}. This idea is central to Relational Frame Theory (RFT), a modern behavioral theory of language and cognition \citep{rft2001, hayes2021relating}. According to RFT, virtually all of human language and higher cognition is founded upon AARR—the ability to treat different stimuli as related along various dimensions (e.g., \textit{same}, \textit{different}, \textit{greater than}, \textit{opposite}, etc.) purely as a result of learned context, not because of any inherent relationship in their physical features.

Three key properties define AARR and distinguish it from simple associative learning:
\begin{enumerate}
    \item \textbf{Mutual Entailment}: This is the bidirectionality of derived relations. If a person learns a relation in one direction (e.g., A is larger than B), they can derive the relation in the opposite direction (B is smaller than A) without direct training \citep{luciano2007role}. In classical terms, mutual entailment encompasses symmetric relations (if $A = B$, then $B = A$) and the inverses of asymmetrical relations (if $A > B$, then $B < A$) in a generalized way. Notably, the derived relation might not be identical in form (for instance, \textit{larger than} vs. \textit{smaller than} are inverse relations rather than exactly the same), but they are mutually implied by each other given the contextual cues (such as the contextual cue for comparison).
    \item \textbf{Combinatorial Entailment}: This is the ability to derive new relations from combinations of learned relations. For example, if one learns that A is related to B, and B is related to C, one can often derive a relation between A and C, depending on the nature of the relation. In the simplest case, if $A = B$ and $B = C$ (coordination relations), then one can derive $A = C$ (equivalence). If $A > B$ and $B > C$ (a comparative relation of “more than”), one can derive $A > C$ (“A is more than C”). These are akin to transitive inferences, but RFT uses the term \textit{combinatorial} entailment to emphasize that the new relation emerges from the combination of two or more other relations.
    \item \textbf{Transformation of Stimulus Function}: Perhaps the most distinctive aspect, this refers to the way the functions of stimuli (their meaning, emotional valence, or behavioral effects) can change based on the relations they participate in \citep{dymond2000transformation}. In other words, if two stimuli are related in a certain way, any psychological function attached to one stimulus (like being pleasant, having a certain name, evoking a specific response) can be transferred to the other stimulus in accordance with their relation. For instance, suppose a person is taught that stimulus A is equivalent to stimulus B ($A = B$, a coordination relation), and separately, stimulus A acquires a particular function (e.g., A is paired with a reward or labeled as “good”). Then, without additional training, the person may treat stimulus B as also having that function (finding B pleasant or “good”), because B is in the same equivalence class as A. If the relation is one of opposition, the functions might transfer in an opposite manner (e.g., if A is opposite to B, and A is associated with “good,” B might be seen as “bad”) \citep{roche2000contextual}. Transformation of function demonstrates how relational learning can govern the meaning of symbols in context.
\end{enumerate}

An example can illustrate these principles. Imagine a scenario in a coffee shop: A newcomer is told that “Espresso is stronger than Americano, and Americano is stronger than Caffé au Lait.” From just this information, the person can derive that Espresso is stronger than Caffé au Lait, and conversely, Caffé au Lait is weaker than Espresso (combinatorial entailment and mutual entailment for the comparative frame). Now, suppose the person actually tastes an Americano and finds it strong and bitter. That experience may attach a function (strong flavor) to Americano. Due to the relational network, the person might now expect that Espresso (which was said to be stronger than Americano) has an even stronger taste, and that Caffé au Lait (weaker than Americano) has a milder taste, even though they have never tasted Espresso or Caffé au Lait. This is a transformation of stimulus function across a comparative relation network: the direct experience with one item (Americano) transformed the anticipated qualities of the related items (Espresso, Caffé au Lait) in line with the learned relations.

Relational Frame Theory has identified numerous types of relational patterns (called \textit{relational frames}) that humans can learn. Some prominent examples include frames of \textit{coordination} (equivalence/sameness), \textit{distinction} (different from), \textit{comparison} (more than/less than as in the coffee strength example), \textit{opposition}, \textit{hierarchy} (e.g., category membership relations, like “X is a kind of Y”), \textit{temporal} (before/after), \textit{spatial} (here/there), and \textit{deictic} (I/you, now/then, here/there, which involve perspective) \citep{rft2001, hayes2021relating}. All these frames share the properties of mutual and combinatorial entailment and can lead to transformations of function, though the exact nature of the entailments depends on the frame.

It is important to note that AARR is considered an \textit{operant behavior}, meaning it is learned through a history of reinforcement and context, rather than being an innate or automatic reflex \citep{hayes2021relating}. The term “arbitrarily applicable” emphasizes that any stimuli, regardless of their formal properties, can be related in any way, given the appropriate training context. Humans, especially those with language ability, seem uniquely capable of this kind of learning. Indeed, research has shown that stimulus equivalence (a basic form of AARR focusing on sameness) reliably appears in humans but not in most non-human animals without language training, with only rare exceptions \citep{schusterman1993}. This link between language and AARR suggests that a capacity for relational responding is a defining feature of higher cognition.

Relational Frame Theory provides a perspective on general intelligence as well. Rather than viewing intelligence as a monolithic IQ or a fixed set of problem-solving abilities, RFT suggests intelligence involves a rich repertoire of relational skills \citep{cassidy2016relational, hayes2021relating}. From this viewpoint, improving one’s ability to learn and manipulate complex relational networks should enhance cognitive performance. Studies have found that training individuals on relational tasks can increase scores on standard intelligence tests \citep{cassidy2016relational}. Programs like \textit{SMART} (Strengthening Mental Abilities with Relational Training) and \textit{PEAK} (Promoting the Emergence of Advanced Knowledge) aim to boost cognitive and language abilities by systematically exercising relational responding abilities \citep{dixon2017teaching}.

In summary, AARR, as characterized by RFT, captures the flexibility, generativity, and context-sensitivity of human symbolic reasoning. Modeling this phenomenon in an AI system requires that the system can represent relations between symbols, infer new relations from old, and dynamically update what symbols mean based on relational context. Next, we discuss NARS, which we propose as a suitable candidate for this challenge.

\subsection{Non-Axiomatic Reasoning System (NARS)}
The Non-Axiomatic Reasoning System (NARS) is an AI system and cognitive architecture developed by Pei Wang \citep{wang2013nalbook, wang2022intelligence} with the goal of realizing a form of general intelligence that operates effectively under real-world constraints. The name “non-axiomatic” reflects that NARS does not assume a predefined, complete set of axioms or truths about the world; instead, it must learn and reason non-axiomatically, meaning all its knowledge is gleaned from experience and is always revisable. NARS was built on the recognition that an intelligent agent in the real world must cope with insufficient knowledge and insufficient resources (a principle Wang abbreviates as AIKR: Assumption of Insufficient Knowledge and Resources; \citep{wang2019defining}). Unlike classical logic systems that are brittle outside of their given axioms, NARS is adaptive and is constantly updating its beliefs and strategies as new information comes in, somewhat akin to a human continually learning and adjusting their understanding.

At the core of NARS is an AI reasoning framework called Non-Axiomatic Logic (NAL). NAL is a formal logic that extends term logic (a kind of logic dealing with relationships between terms or concepts) and is probabilistic in nature. NARS uses an internal language, Narsese, to represent knowledge. All pieces of knowledge in NARS are expressed as statements in Narsese, which typically have a subject and a predicate and a copula connecting them (the copula defines the type of relation between subject and predicate). The simplest form is an inheritance relation “$S \to P$” meaning “$S$ is a kind of $P$” or “$S$ implies $P$” in a category sense. For example, one could represent “Tweety is a bird” as $\texttt{Tweety} \to \texttt{Bird}$, and “Birds are animals” as $\texttt{Bird} \to \texttt{Animal}$. NAL can then derive $\texttt{Tweety} \to \texttt{Animal}$ by inference (a kind of syllogism) \citep{wang2013nalbook}. In addition to inheritance, Narsese includes other basic copulas such as similarity (noted as $\leftrightarrow$ in Narsese, meaning two terms are similar or equivalent in some sense), implication ($\rightarrow$ with different context indicating temporal or causal implication), and equivalence ($\Leftrightarrow$ for bi-conditional statements). Through combinations of these, NARS can represent a wide variety of knowledge, including rules like “if X happens then Y tends to happen” (an implication), or “Concept A is similar to Concept B” (a similarity statement).

Crucially, every statement in NARS carries a measure of uncertainty. NARS does not use binary true/false assignments; instead, each piece of knowledge has a truth value with two parameters: \textit{frequency} (a measure akin to probability based on how often the relation has been true in experience) and \textit{confidence} (reflecting the amount of evidence available) \citep{hammer2022reasoning, wang2022intelligence}. This allows NARS to reason under uncertainty and update its beliefs as new evidence arrives. For example, if initially NARS has little evidence about “Tweety can fly,” it might assign it a low confidence. If many observations confirm it, the confidence (and perhaps the frequency) increases.

Another distinguishing feature of NARS is its approach to resource constraints. NARS operates in real-time and has a limited “budget” for attention and memory. It cannot consider all knowledge all the time. Instead, it uses a priority mechanism to decide which tasks (questions, goals, new knowledge) to process next, based on factors like urgency and relevance. This ensures that at any given moment, the system focuses on the most pertinent information, allowing it to scale to larger problems by not getting bogged down in less relevant details.

Recent implementations of NARS include OpenNARS and specifically a variant called OpenNARS for Applications (ONA) \citep{hammer2020opennars}. ONA is tailored for integration into practical applications, including robotics. It extends the basic NARS framework with sensorimotor capabilities, meaning it can handle input from sensors and send output to actuators (motors) as part of its reasoning. This is done by treating sensorimotor events also as terms in the language (for instance, a sensory observation or a motor command can be a term that participates in statements). In ONA, the reasoning engine is capable of doing temporal inference, understanding sequences of events and causality. Temporal relations in Narsese might be represented with additional notation - for example, $A =/> B$ might denote events $A$ and $B$ happening in sequence. ONA’s design includes components like event buffers, concept memory, and distinct inference processes for different types of tasks (e.g., some for immediate reactions, some for long-term learning) \citep{hammer2020opennars, hammer2022reasoning}.

For the purposes of this work, what is important is that NARS (and ONA) provides:
\begin{itemize}
    \item A flexible knowledge representation that can express arbitrary relations between symbols (via terms and copulas in Narsese).
    \item Inference rules that can derive new relationships from known ones, analogous to the entailments described in RFT. For example, NARS can perform syllogistic inference (if $A \to B$ and $B \to C$, derive $A \to C$) and inductive inference (generalizing or specializing relations based on evidence), which parallel combinatorial entailment in AARR.
    \item The ability to incorporate new knowledge on the fly and revise existing knowledge, which is essential for any learning system attempting to acquire relational behavior through training.
    \item The ability to handle context and switch between tasks, somewhat akin to how contextual cues in AARR determine which relation applies. In NARS, context is handled through its concept activations and the specific questions posed to the system; it is not identical to the notion of contextual cues in RFT, but NARS can take context into account by treating it as just another piece of information in the premise of a statement or rule.
\end{itemize}

In short, NARS can be seen as a unified cognitive model that does not separate reasoning, learning, memory, and perception into different modules; the same underlying logic and control mechanism handles all these functions \citep{wang2022unified}. This makes it very appealing for modeling complex cognitive phenomena like AARR, because we do not need to bolt together separate systems for learning relations and for reasoning about them — NARS does both in one framework. The challenge is to design the right way to present relational training to NARS and possibly to extend NARS with any additional mechanisms so that it can exhibit mutual and combinatorial entailment and transformation of functions in a manner comparable to humans.

\subsection{Machine Psychology}

Machine Psychology is an interdisciplinary framework that integrates learning psychology with an adaptive AI system, the Non-Axiomatic Reasoning System (NARS), to explore the emergence of cognitive behaviors in artificial agents \citep{johansson2024empirical, johansson2024machine}. This approach systematically investigates increasingly complex learning processes, drawing from operant conditioning, generalized identity matching, and functional equivalence, which are fundamental to relational cognition. In Table \ref{table_empirical_studies}, we clarify how this systematic approach has been carried out in previous studies.

In this work, we assume that the system is interacting with the environment using different sensors. A key sensor that will be used throughout the entire paper is the assumption of a location sensor. Objects perceived by the vision system would using this model all be assigned a location. The labels \texttt{sample}, \texttt{left}, \texttt{right}, etc, are totally arbitrary. They are chosen by the designer and are only labels used to indicate that different objects are perceived at different locations. 

\begin{figure}
\centering
\includegraphics[scale=0.15]{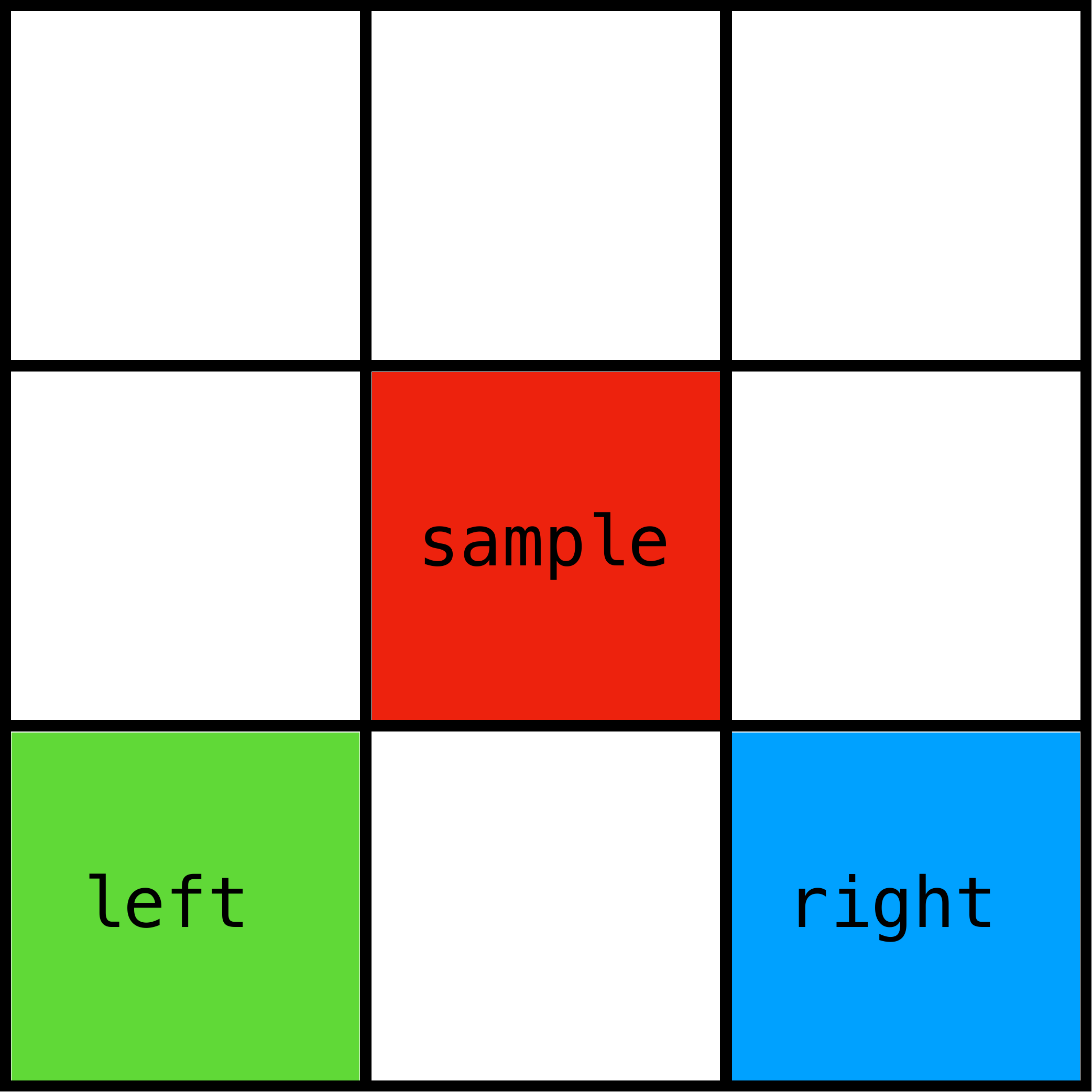}
\caption{
An example scene where the system perceives three different colors at three different locations.
} 
\label{fig_mts_colors_left_right}
\end{figure}

We could also imagine that the system is equipped with a color sensor, and is interacting with a Matching-to-sample procedure. For example, as illustrated in Figure \ref{fig_mts_colors_left_right}, something red is in the sample position, something green is to the left, and something blue to the right. This could be described that the only “eyes” that the system have are location and color, meaning that other object properties like shape and size couldn't be perceived by that system.

The way we represent such interactions with the world in this paper is like the following:

\begin{lstlisting}[language=c]
<(sample * red) --> (loc * color)>. :|:
<(left * green) --> (loc * color)>. :|:
<(right * blue) --> (loc * color)>. :|:
\end{lstlisting}

The scene is described by two temporal statements (as indicated by \texttt{:|:}). Perceiving a green object to the left can be described as an interaction between perceiving to the left, and perceiving green. Hence, the statement \texttt{<(left * green) -{}-> (loc * color)>} can be seen as a composition of \texttt{<left -{}-> loc>} and \texttt{<green -{}-> color>}. This encoding of object properties at certain locations will be used throughout this paper. Importantly, also an OCR detector will be assumed in the experiments carried out in the present study.

\subsubsection{Operant Conditioning}

The foundation of Machine Psychology is built on operant conditioning, a fundamental mechanism of adaptive behavior \citep{johansson2024machine}. In our research, NARS was exposed to operant contingencies where behaviors were reinforced based on temporal and procedural reasoning. This enabled NARS to learn through interaction with its environment, adjusting actions based on feedback, similar to how organisms learn in response to consequences. The results demonstrated that NARS could acquire and refine behaviors through reinforcement, providing an essential basis for more advanced relational learning.

\begin{lstlisting}[language=c]
<(left * blue) --> (loc * color)>. :|:
<(right * green) --> (loc * color)>. :|:
G! :|:

// Executed with motor babbling:
// ^select executed with args ({SELF} * right)

G. :|:

// Derived with frequency 1, and confidence 0.19:
// <(<(right * green) --> (loc * color)> &/ <({SELF} * right) --> ^select>) =/> G>.
\end{lstlisting}

\subsubsection{Generalized Identity Matching}

Building upon operant conditioning, our research extended into generalized identity matching, which involves recognizing and responding to identity relations across varying stimuli \citep{johansson2023generalized}. This required NARS to utilize complex learning mechanisms, including abstraction and relational generalization. By introducing an abstraction mechanism to NARS, we enabled it to derive identity relations beyond explicit training examples, mirroring human cognitive abilities in symbolic matching tasks. The results showed that NARS could generalize identity relations to novel stimuli, demonstrating an emergent form of relational reasoning.

Let's say that the system was exposed to the following NARS statements in the training phase:

\begin{lstlisting}[language=c]
<(sample * blue) --> (loc * color)>. :|:
<(left * green) --> (loc * color)>. :|:
<(right * blue) --> (loc * color)>. :|:
G! :|:
\end{lstlisting}

NARS could execute \texttt{\^{}match} with \texttt{sample} and \texttt{right} (from motor babbling or a decision based on previous experience), which would be considered correct, and hence the feedback \texttt{G. :|:} would be given to NARS, followed by 100 time steps. Only from this single interaction, NARS would form both a specific and a general hypothesis:

\begin{lstlisting}[language=c]
<((<(sample * blue) --> (loc * color)> &/ 
 <(right * blue) --> (loc * color)>) &/ 
 <({SELF} * (sample * right)) --> ^match>) =/> G>
// frequency: 1.00, confidence: 0.15

<((<(#1 * #2) --> (loc * color)> &/ 
 <(#3 * #2) --> (loc * color)>) &/ 
 <({SELF} * (#1 * #3)) --> ^match>) =/> G>
// frequency: 1.00, confidence: 0.15
\end{lstlisting}

\subsubsection{Functional Equivalence}

Further advancing Machine Psychology, we explored functional equivalence, a process in which stimuli become interchangeable in guiding behavior due to shared functional properties \citep{johansson2024functional}. This study introduced additional inference mechanisms into NARS, allowing it to derive new relations based on implications and acquired equivalences. Functional equivalence is critical for understanding how abstract categories are formed and used in problem-solving. Our findings indicate that NARS can establish and apply functional equivalence relations, effectively transferring learned functions between distinct but related stimuli.

\begin{lstlisting}[language=c]
<(s1 * A1) --> (loc * ocr)>. :|:
G! :|:

// Executed with motor babbling
<({SELF} * R1) --> ^press>. :|:

G. :|:

// Derived
<(<(s1 * A1) --> (loc * ocr)> &/ 
 <({SELF} * R1) --> ^press>) =/> G>.

100

<(s1 * A2) --> (loc * ocr)>. :|:
G! :|:

// Executed same operation with motor babbling
<({SELF} * R1) --> ^press>. :|:

G. :|:

// Derived
<(<(s1 * A2) --> (loc * ocr)> &/ 
 <({SELF} * R1) --> ^press>) =/> G>.
\end{lstlisting}

Since the system derived two contingencies that only differed in the pre-condition, statements like the following (functional equivalence) would also be derived:

\begin{lstlisting}[language=c]
<<($1 * A1) --> (loc * ocr)> ==> <($1 * B1) --> (loc * ocr)>>.
<<($1 * B1) --> (loc * ocr)> ==> <($1 * A1) --> (loc * ocr)>>.
\end{lstlisting}

These studies collectively illustrate the progression from simple operant conditioning to complex relational cognition, reinforcing Machine Psychology as a viable framework for advancing artificial general intelligence (AGI). An overview of the systematic approach Machine Psychology has taken, can be seen in Table \ref{table_empirical_studies}. By systematically integrating behavioral learning principles with adaptive AI reasoning, this approach contributes to the development of more flexible, human-like intelligence in machines.

\section{Related Work}
Integrating principles of human cognition and learning into AI systems is a growing interdisciplinary endeavor. However, Relational Frame Theory (RFT) and its core concept of AARR have seen relatively little application in mainstream AI research. Most approaches to relational reasoning in AI have taken different paths:

\begin{itemize}
    \item \textbf{Symbolic AI \& Knowledge Graphs}: Traditional symbolic reasoning systems (e.g., knowledge graph inference engines or logic-based AI) typically handle relations between symbols, but these relations are usually \textit{axiomatically defined} \citep{lenat1995cyc, rosenbloom2016sigma}. For example, ontologies explicitly define inverse relations (“isFatherOf” as the inverse of “isChildOf”) or symmetry (“sibling” relations). Such systems generally do not \textit{learn} these relations; instead, they depend on predefined knowledge structures. In contrast, AARR emphasizes the ability to \textit{learn arbitrary} relations from experience and context, dynamically deriving novel conclusions. Our approach makes use of NARS precisely to achieve relational learning from interaction rather than relying on static ontological axioms.

    \item \textbf{Machine Learning for Relational Tasks}: In machine learning, approaches like \textit{relational reinforcement learning}, \textit{graph neural networks}, and \textit{transformer-based models} capture patterns in relational data effectively. DeepMind’s Relation Networks, for example, can learn relational properties from large datasets, answering questions like “Is object A above object B?” in images \citep{santoro2017simple}. Despite their power, these models typically require substantial training data and do not explicitly guarantee properties like mutual or combinatorial entailment; these properties must implicitly emerge, often requiring explicit and extensive training. Additionally, neural models frequently lack interpretability, unlike logic-based systems like NARS, and tend to struggle with one-shot or few-shot relational generalization — a hallmark of human cognition that our NARS-based approach explicitly seeks to replicate.

    \item \textbf{Bayesian Approaches to Relational Learning}: Bayesian methods such as probabilistic programming and Bayesian relational modeling explicitly represent uncertainty, allowing for principled relational inference \citep{nitti2016probabilistic,tenenbaum2006theory}. These approaches excel at generalizing from sparse data but often depend heavily on predefined model structures and priors. Consequently, they typically cannot dynamically derive novel relational structures purely from experience or flexibly model context-sensitive entailments. In contrast, our NARS-based method dynamically constructs relational structures directly from experience and explicitly captures relational entailment and transformation of stimulus functions without relying on strong initial assumptions.
\end{itemize}

Regarding computational approaches that have explicitly modeled AARR, very few examples exist. Early computational models in the 1990s and early 2000s attempted to simulate \textit{derived stimulus relations} such as stimulus equivalence using connectionist neural networks, treating relational responding as an activation pattern to be learned \citep{barnes1993stimulus, cullinan1994transfer}. While these models could demonstrate symmetry and transitivity under certain conditions, they typically required extensive and careful exemplar training and had difficulties scaling to complex relational networks. Computational modeling of stimulus equivalence is an active area of research \citep{tovar2023computational}, but approaches inspired explicitly by RFT that extend beyond stimulus equivalence remain rare. To our knowledge, no prior published work has demonstrated a computational approach to AARR capable of flexibly scaling to multiple and diverse relational structures (but see the work of \cite{edwards2022functional, edwards2024functional}).

In summary, while relational reasoning broadly remains a vibrant area of AI research, the unique challenge of \textit{learning arbitrary contextual relations and deriving untrained relations} — the hallmark of human relational flexibility — remains largely unmet. By making use of NARS, our approach directly addresses this gap. To the best of our knowledge, our work represents the first demonstration of a reasoning system explicitly capturing \textit{mutual entailment, combinatorial entailment, and transformation of function} within a unified computational framework. It therefore establishes foundational groundwork for future AI systems capable of learning and reasoning more like humans—not by mimicking neural architectures but by explicitly implementing the functional principles underlying human cognition.

\section{Modeling AARR in NARS}

To enable the modeling of Arbitrarily Applicable Relational Responding (AARR) within OpenNARS for Applications (ONA), we introduce a novel mechanism called \textit{acquired relations}. Currently, ONA's reasoning is based primarily on sensorimotor contingencies; however, according to NARS theory (NAL Definition 8.1 in \cite{wang2013nalbook}), relational terms (\textit{products}) can equivalently be represented as compound terms of inheritance statements. This theoretical notion has not yet been implemented in ONA, and its introduction would allow the system to explicitly derive relational statements directly from learned sensorimotor contingencies.

Within NARS theory, a learned contingency such as:
\begin{lstlisting}[language=c]
<((<A1 --> p1> &/ <B1 --> q1>) &/ ^left) =/> G>.
\end{lstlisting}

can yield an \textit{acquired relation}, formally represented as:
\begin{lstlisting}[language=c]
<(A1 * B1) --> (p1 * q1)>.
\end{lstlisting}

In the notation employed here, learned sensorimotor contingencies often take the form:
\begin{lstlisting}[language=c]
<(sample * red) --> (loc * color)> &/ 
 <(left * blue) --> (loc * color)> &/ 
 <({SELF} * (sample * left)) --> ^match> =/> G>.
\end{lstlisting}

Following our approach, this yields two distinct relational terms — one describing the relation between stimulus properties (colors), and another describing the relational structure of stimulus locations:
\begin{lstlisting}[language=c]
<(red * blue) --> (color * color)> && 
 <(sample * left) --> (loc * loc)>
\end{lstlisting}

To avoid a \textit{combinatorial explosion}, i.e., an exponential growth in derived terms and inferences, the introduction of acquired relations is carefully restricted. Specifically, new relations are generated only when procedural operations within contingencies are actively executed by the system. This targeted triggering ensures computational efficiency while maintaining functional generality.

Acquired relations can be combined with \textit{implications}, another core element in NARS theory (see statement-level inference in \cite{wang2013nalbook}), allowing for generalized, context-sensitive reasoning. For example, from the acquired relations shown previously, the following implications can be derived:

\begin{lstlisting}[language=c]
<(red * blue) --> (color * color)> && 
 <(sample * left) --> (loc * loc)> ==>
 <(sample * red) --> (loc * color)> &/ 
  <(left * blue) --> (loc * color)> &/ 
  <({SELF} * (sample * left)) --> ^match> =/> G>.
\end{lstlisting}

More generally, implications abstracted with variables take this form:

\begin{lstlisting}[language=c]
<($1 * $2) --> (color * color)> && 
 <($3 * $4) --> (loc * loc)> ==>
 <($3 * $1) --> (loc * color)> &/ 
  <($4 * $2) --> (loc * color)> &/ 
  <({SELF} * ($3 * $4)) --> ^match> =/> G>.
\end{lstlisting}

This framework can be understood as a \textit{grounding mechanism} whereby abstract relations (e.g., color-color) become anchored in concrete sensorimotor experiences. This allows NARS to dynamically transition from basic, animal-like contingency learning towards symbolic, human-like reasoning capabilities.

Conversely, symbolic-level relational statements can also guide sensorimotor behavior. If a relation such as $(blue \rightarrow yellow)$ is symbolically derived, it can then inform decision-making in novel situations via the implications described above, provided relevant locational relations (e.g., $(sample \rightarrow right)$) are established through direct interaction with the environment.

The concept of acquired relations is general and not restricted to matching-to-sample procedures. For example, functional equivalences acquired through interactions with different procedures also lead to relational derivations. Consider the following example:

\begin{lstlisting}[language=c]
<(<(left * green) --> (loc * color)> &/ 
  <({SELF} * left) --> ^select>) =/> G>

100 // Wait 100 time steps

<(<(left * blue) --> (loc * color)> &/ 
  <({SELF} * left) --> ^select>) =/> G>

// Derived functional equivalence:
<(left * green) --> (loc * color)> <=> 
 <(left * blue) --> (loc * color)>
\end{lstlisting}

This equivalence, in turn, can support acquired relational implications:

\begin{lstlisting}[language=c]
<(green * blue) --> (color * color)> && 
 <(left * left) --> (loc * loc)> ==>
 <(left * green) --> (loc * color)> <=> 
  <(left * blue) --> (loc * color)>

// Abstracted form:
<($1 * $2) --> (color * color)> && 
 <($3 * $3) --> (loc * loc)> ==>
 <($3 * $1) --> (loc * color)> <=> 
  <($3 * $2) --> (loc * color)>
\end{lstlisting}

This flexibility aligns closely with contemporary learning psychology perspectives, which argue that any regularity — such as stimulus pairing or common roles within contingencies — can serve as a contextual cue for relational responding \citep{dehouwer2020book, hughes2016intersecting}.

In the following section, we detail specific experimental paradigms designed to validate and explore the capabilities enabled by these modeling extensions.

\section{Methods}
\begin{figure}[h!]
\begin{center}
\includegraphics[scale=0.75]{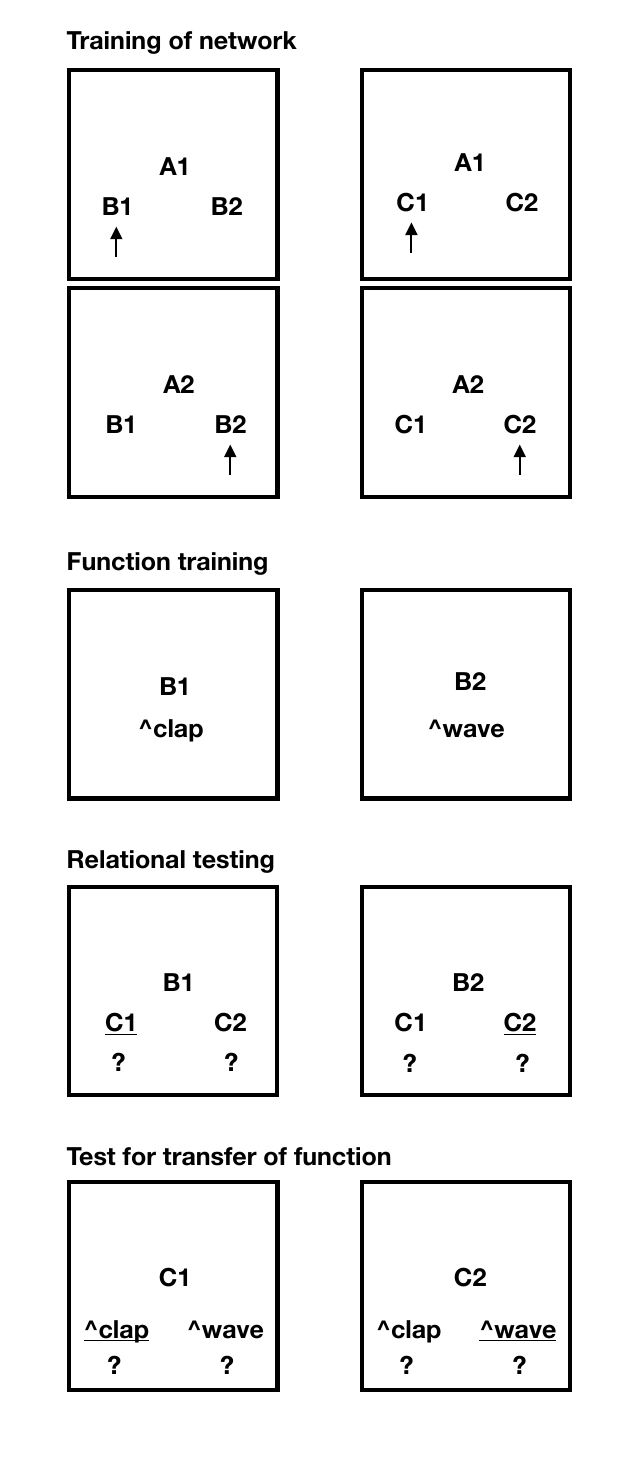}
\end{center}
\caption{
Task 1 of this paper. Stimulus equivalence and the transfer of function. The necessary pre-training (Phase 1) is excluded from the picture. Picture shows Phases 2-5 of the task. Underlined options indicate correct choices.
} 
\label{fig_aarr_tasks1}
\end{figure}

\begin{figure}[h!]
\begin{center}
\includegraphics[scale=0.6]{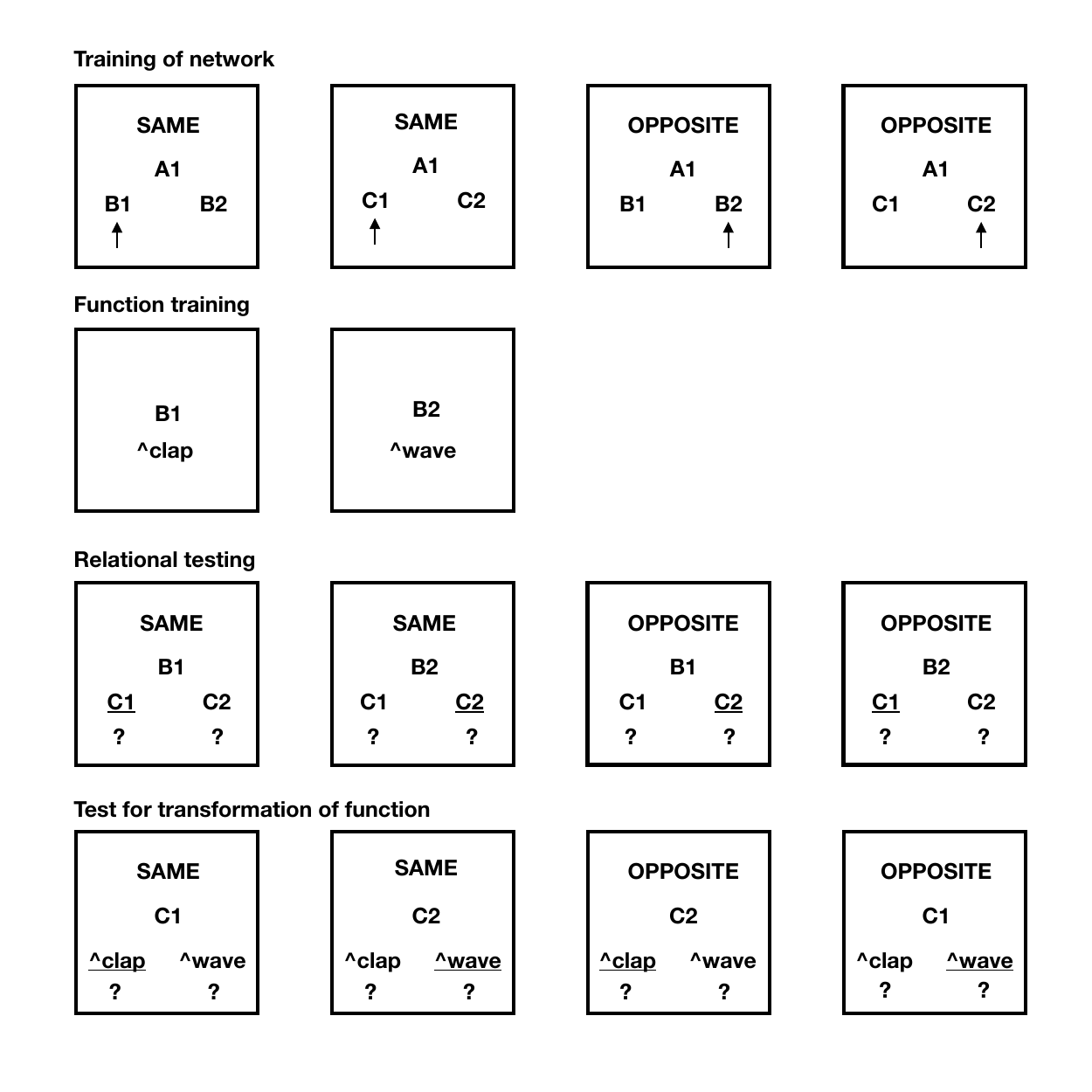}
\end{center}
\caption{
Task 2 of this paper. AARR in accordance with opposition and the transformation of function. The necessary pre-training (Phase 1) is excluded from the picture. Picture shows Phases 2-5 of the task. Underlined options indicate correct choices.} 
\label{fig_aarr_tasks2}
\end{figure}

The experiments described here are theoretical demonstrations designed to show that the NARS logic, when extended with the enhancements proposed earlier (i.e., acquired relations and implications), is capable of modeling Arbitrarily Applicable Relational Responding (AARR). Specifically, we designed two experiments adapted from human studies commonly reported in the Relational Frame Theory (RFT) literature: the Stimulus Equivalence and Function Transfer task (Task 1; Figure \ref{fig_aarr_tasks1}) and the Opposition and Function Transformation task (Task 2; Figure \ref{fig_aarr_tasks2}) \citep{hayes1987stimulus, roche2000contextual}. These tasks were modified to suit the capabilities of the NARS framework. Importantly, the described experimental setups were not implemented in the existing OpenNARS for Applications (ONA) system \citep{hammer2020opennars}; rather, they are presented here as conceptual analyses employing symbolic representations exclusively (no physical robots were involved) to clearly illustrate how NARS, once extended, could account for these forms of relational reasoning.

\subsection{Task 1: Stimulus Equivalence and Transfer of Function}

The design for Task 1 was inspired by the methodology introduced by \citet{hayes1987stimulus}. In their original human study, participants underwent four phases: (1) training conditional discriminations, (2) testing for derived equivalence classes, (3) training discriminative stimulus functions on selected class members, and (4) testing whether discriminative functions transferred to other members of the same equivalence classes. Importantly, the original study did not explicitly account for participants' prior relational learning history.

In the present study, we explicitly included pretraining to establish basic relational skills prior to the main experiments. The study consisted of four phases conducted sequentially:

\begin{enumerate}
	\item \textbf{Pretraining of relational networks: } This phase explicitly trained foundational relations such as symmetry ($X1 \rightarrow Y1$ and $Y1 \rightarrow X1$), and transitivity ($X1 \rightarrow Y1$, $Y1 \rightarrow Z1$, thus deriving $X1 \rightarrow Z1$).
		
	\item \textbf{Training conditional discriminations: } Using a Matching-to-sample (MTS) procedure, conditional discriminations were trained within two separate stimulus networks: one comprising stimuli $A1$, $B1$, and $C1$, and another comprising $A2$, $B2$, and $C2$.
	
	\item \textbf{Function training: } NARS was trained to execute two discriminative responses: \verb!^clap! when $B1$ was presented as a sample stimulus, and \verb!^wave! when $B2$ appeared as the sample.
	
	\item \textbf{Testing derived relations and transfer: } In the final phase, derived relations within each $ABC$ network were tested without feedback, specifically examining whether previously trained discriminative functions (\verb!^clap!, \verb!^wave!) transferred to equivalent stimuli ($C1$, $C2$).
\end{enumerate}

\subsection{Task 2: Opposition and Transformation of Function}

Task 2 was inspired by the relational methodology of \citet{roche2000contextual}. Roche and colleagues examined how derived relational responses and stimulus functions transformed contextually using “Same” and “Opposite” relational frames. Their human participants initially learned operant associations between arbitrary stimuli and actions (e.g., waving, clapping), followed by relational pretraining to establish “Same” and “Opposite” frames. Through training and contextual cueing, participants showed contextually controlled derived responding (e.g., relationally responding “Same” or “Opposite” for specific stimuli) and function transformation.

In the current study, we again included explicit pretraining phases to equip NARS with necessary relational skills. The experimental design comprised five phases:

\begin{enumerate}
	\item \textbf{Pretraining of relational frames: } This phase explicitly trained “SAME” and “OPPOSITE” relations, establishing mutual entailment (e.g., SAME $X1 \leftrightarrow Y1$, OPPOSITE $X1 \leftrightarrow Y2$) and combinatorial entailment (e.g., SAME $X1 \rightarrow Y1$, SAME $Y1 \rightarrow Z1$, thus deriving SAME $X1 \rightarrow Z1$). Functional equivalence and explicit transfers between symmetry and functional equivalence were also established.
		
	\item \textbf{Training relational networks: } Using the Matching-to-sample (MTS) procedure, relational networks were trained, forming explicit SAME (e.g., $A1 \rightarrow B1$, $A1 \rightarrow C1$) and OPPOSITE ($A1 \rightarrow B2$, $A1 \rightarrow C2$) relations. A second analogous network ($A2$-$B2$-$C2$) was similarly trained.
	
	\item \textbf{Function training: } The system was trained to produce discriminative responses \verb!^clap! (for $B1$) and \verb!^wave! (for $B2$).
	
	\item \textbf{Testing derived relations and function transformations: } In the final phase, derived relations within the SAME/OPPOSITE networks were tested without feedback, specifically examining whether trained functions transformed appropriately across relational contexts. Stimuli tested included combinations such as SAME/$C1$, SAME/$C2$, OPPOSITE/$C1$, and OPPOSITE/$C2$.
\end{enumerate}

\section{Results}

Here, we summarize the key outcomes of our theoretical demonstrations evaluating whether NARS, with the proposed extensions, can model Arbitrarily Applicable Relational Responding (AARR). Detailed step-by-step training, derivation processes, and extended test examples can be found in the Supplementary Material.

\subsection{Stimulus Equivalence and Transfer of Function}

In the first experiment (illustrated in Figure \ref{fig_aarr_tasks1}), we explored whether NARS logic could model the formation of stimulus equivalence classes and demonstrate the transfer of stimulus functions across related stimuli. Briefly, NARS was theoretically exposed to matching-to-sample (MTS) procedures where conditional relations ($A \rightarrow B$ and $B \rightarrow C$) were explicitly trained. Additionally, discriminative functions were assigned to specific stimuli within these relational networks (e.g., stimulus $B_1$ triggering a \verb!^clap! response, and $B_2$ a \verb!^wave! response).

Key results included:

\begin{itemize}
    \item \textbf{Mutual entailment:} NARS successfully derived bidirectional relations (e.g., if trained $A \rightarrow B$, it inferred $B \rightarrow A$).
    \item \textbf{Combinatorial entailment:} The system correctly inferred indirect relations from explicitly trained ones (e.g., from $A \rightarrow B$ and $B \rightarrow C$, it inferred $A \rightarrow C$).
    \item \textbf{Transformation of function:} Critically, discriminative functions (e.g., \verb!^clap! and \verb!^wave!) initially trained on $B$-stimuli were transferred without additional training to $C$-stimuli through derived equivalence relations, demonstrating a successful relational transfer of stimulus functions.
\end{itemize}

Thus, NARS logic adequately models essential aspects of stimulus equivalence and function transfer, foundational within Relational Frame Theory (Figure \ref{fig_abc_same}; detailed derivations in Supplementary Material Section 1).

\subsection{Opposition and Transformation of Function}

In the second experiment (illustrated in Figure \ref{fig_aarr_tasks2}), we assessed whether NARS logic could model relational networks involving oppositional frames (“SAME” and “OPPOSITE”) and the contextual transformation of stimulus functions. Similar to the first task, MTS training was theoretically applied, but now relations explicitly involved both SAME and OPPOSITE contexts. After training, discriminative functions were again assigned to specific stimuli within these networks.

Key outcomes included:

\begin{itemize}
    \item \textbf{Context-sensitive mutual entailment and combinatorial entailment:} NARS derived relations consistent with trained SAME and OPPOSITE relational frames, correctly generalizing from explicitly trained examples.
    \item \textbf{Transformation of function across oppositional relations:} Trained discriminative functions (e.g., \verb!^clap! associated with stimulus $B_1$, and \verb!^wave! with $B_2$) were accurately transferred to related stimuli ($C_1$ and $C_2$), including appropriate reversal in functions when oppositional relational contexts were applied (e.g., if stimulus pairs were related as OPPOSITE, stimulus functions reversed accordingly).
\end{itemize}

These results illustrate that NARS logic effectively models complex, contextually controlled transformations of function, consistent with Relational Frame Theory (Figure \ref{fig_abc_opposite}; detailed derivations in Supplementary Material Section 2).

In summary, these theoretical demonstrations confirm that the extended NARS logic is sufficiently powerful and flexible to capture core relational learning phenomena — mutual entailment, combinatorial entailment, and transformation of function — essential for modeling human-like symbolic reasoning and cognition.

\section{Discussion and Conclusion}

This study demonstrated that the Non-Axiomatic Reasoning System (NARS), extended with mechanisms inspired by Relational Frame Theory (RFT), can successfully model Arbitrarily Applicable Relational Responding (AARR), a cornerstone of human cognition. Through theoretical analysis and logical derivations, we showed how NARS’s adaptive logic can capture essential relational learning phenomena without pre-defined axioms or extensive data-driven training. This integration provides a computational framework aligning cognitive science principles with artificial intelligence (AI), underscoring the interdisciplinary potential of Machine Psychology \citep{johansson2024empirical, johansson2024machine} in developing flexible, context-sensitive reasoning systems.

\subsection{Summary of Findings}

We have shown theoretically that NARS can replicate critical aspects of human-like relational reasoning by modeling Arbitrarily Applicable Relational Responding. Specifically, we demonstrated that:

\begin{itemize}
    \item NARS exhibits \textit{mutual entailment}, accurately deriving bidirectional relations from trained unidirectional associations.
    \item It demonstrates robust \textit{combinatorial entailment}, integrating multiple explicitly trained relations to correctly infer novel relations.
    \item It successfully replicates \textit{transformation of stimulus function}, whereby functions (such as specific responses like “clap” or “wave”) trained to one stimulus are systematically transferred to other related stimuli without additional direct training.
\end{itemize}

These findings illustrate that the cognitive mechanisms underlying AARR — once considered unique to biologically evolved cognition — can be conceptually instantiated within a symbolic reasoning system. NARS’s capability to learn from minimal, structured experiences and subsequently perform flexible relational inference provides a clear departure from contemporary AI models that primarily rely on large-scale statistical training. Instead, our approach emphasizes “small data” and logical consistency, aligning closely with the RFT premise that very few exemplars, combined with appropriate contextual cues, can generate powerful relational generalizations.

\subsection{Implications for Artificial General Intelligence}

Our theoretical demonstration of AARR within NARS offers significant implications for AGI research. First, it illustrates that sophisticated relational reasoning is achievable through adaptive symbolic systems without relying on extensive datasets, reinforcing structured symbolic learning as a viable path toward AGI. Second, our approach establishes learning psychology principles — particularly those articulated by RFT — as functional benchmarks for evaluating AGI systems’ relational generalization capabilities. Third, the flexibility of NARS in dynamically constructing relational structures under uncertainty makes it suitable for adaptive, real-world contexts. Lastly, integrating adaptive logic with relational reasoning supports broad applications, including robotics and human-AI interaction, where context-sensitive symbolic manipulation is essential for achieving human-like understanding.

\subsection{Future Directions}

This theoretical study opens several avenues for further exploration. One immediate direction involves expanding the relational frames modeled in NARS beyond equivalence and opposition, including comparative, hierarchical, and deictic relations, to comprehensively evaluate the system’s generalization capabilities. Another promising direction involves scaling relational networks by increasing stimuli complexity, testing NARS’s resource management and inference flexibility. Additionally, integrating perceptual inputs with symbolic reasoning represents a crucial step toward practical, embodied applications, enabling NARS to generate and reason about relations directly from sensory data in dynamic environments. Lastly, further refining and automating the relational learning mechanisms within NARS, alongside comparisons of NARS-derived relational learning curves with empirical human data, could guide targeted enhancements and deepen our understanding of relational cognition in both artificial and biological systems.

\subsection{Conclusion}

We presented a theoretical framework demonstrating that NARS, enhanced by relational learning principles derived from Relational Frame Theory, can successfully model Arbitrarily Applicable Relational Responding — a foundational component of human cognition. This provides a concrete method for developing symbolic AI systems capable of dynamic, context-sensitive relational reasoning similar to that observed in humans. These findings represent a meaningful step toward bridging cognitive science and artificial intelligence, emphasizing that principles identified through human learning research can inform AI systems that “think” more like humans — not necessarily in brain-like structures but in the dynamic and contextually controlled use of symbolic knowledge. Continued interdisciplinary research in this direction holds considerable promise for developing flexible, adaptive, and ultimately more human-like artificial intelligence.

\section*{Conflict of Interest Statement}

The authors declare that the research was conducted in the absence of any commercial or financial relationships that could be construed as a potential conflict of interest.

\section*{Author Contributions}

RJ solely conceived and designed the study, performed the experiments, analyzed the data, and wrote the manuscript.


\section*{Funding}
The author(s) declare financial support was received for the research, authorship, and/or publication of this article. This work was in part financially supported by Digital Futures through grant agreement KTH-RPROJ-0146472.


\section*{Acknowledgments}
The author would like to thank Patrick Hammer and Tony Lofthouse for many valuable discussions regarding the work presented in this paper.





\bibliographystyle{apalike}
\bibliography{aarr.bib}  

\section*{Supplementary Material}
\setcounter{section}{0}
\renewcommand{\thesection}{S\arabic{section}}
\setcounter{subsection}{0}
\renewcommand{\thesubsection}{S\arabic{section}.\arabic{subsection}}

\section{Stimulus Equivalence and Transfer of Function}

\subsection{Phase 1: Pretraining of relational networks}

\subsubsection{Learning Conditionality} \label{paragraph:learn_cond}

First, the $X1 \rightarrow Y1$ relation was trained using the Matching-to-sample procedure:

\begin{lstlisting}[language=c]
<(sample * X1) --> (loc * ocr)>. :|:
<(left * Y1) --> (loc * ocr)>. :|:
<(right * Y2) --> (loc * ocr)>. :|:
G! :|:
\end{lstlisting}

Motor babbling would be triggered, and with feedback, \texttt{G} would be provided as a consequence when \texttt{<(\{SELF\} * (sample * left)) -{}-> \^{}match>} was executed.

After execution, the following contingency would be derived:

\begin{lstlisting}[language=c]
<(sample * X1) --> (loc * ocr)> &/ <(left * Y1) --> (loc * ocr)> &/ <({SELF} * (sample * left)) --> ^match>) =/> G>.
\end{lstlisting}

Since the system acted on this contingency, the following two relations would be acquired:

\begin{lstlisting}[language=c]
<(X1 * Y1) --> (ocr * ocr)> && <(sample * left) --> (loc * loc)>.
\end{lstlisting}

With these derived, an implication between the acquired $X1 \rightarrow Y1$ and $sample \rightarrow left$ relations and the corresponding contingency would also be derived:

\begin{lstlisting}[language=c]
<(X1 * Y1) --> (ocr * ocr)> && <(sample * left) --> (loc * loc)>
 ==>
<(sample * X1) --> (loc * ocr)> &/ <(left * Y1) --> (loc * ocr)> 
 &/ <({SELF} * (sample * left)) --> ^match>) =/> G>
\end{lstlisting}

Finally, a general form of this implication, with variables introduced, would also be established:

\begin{lstlisting}[language=c]
<($1 * $2) --> (ocr * ocr)> && <($3 * $4) --> (loc * loc)>
 ==>
<($3 * $1) --> (loc * ocr)> &/ <($4 * $2) --> (loc * ocr)> &/ 
 <({SELF} * ($3 * $4)) --> ^match>) =/> G>
\end{lstlisting}

\subsubsection{Learning Symmetry}

After the $X1 \rightarrow Y1$ relation has been trained, the system could be exposed to a matching-to-sample situation where the $Y1 \rightarrow X1$ could be trained:

\begin{lstlisting}[language=c]
<(sample * Y1) --> (loc * ocr)>. :|:
<(left * X1) --> (loc * ocr)>. :|:
<(right * X2) --> (loc * ocr)>. :|:
G! :|:
\end{lstlisting}

With similar learning as in the last paragraph, the system would acquire the following:

\begin{lstlisting}[language=c]
<(Y1 * X1) --> (ocr * ocr)> && <(sample * left) --> (loc * loc)>
 ==>
<(sample * Y1) --> (loc * ocr)> &/ <(left * X1) --> (loc * ocr)> 
 &/ <({SELF} * (sample * left)) --> ^match>) =/> G>
\end{lstlisting}

Symmetry, represented with a functional equivalence, could then be derived using the acquired $X1 \rightarrow Y1$ and $Y1 \rightarrow X1$ relations.

\begin{lstlisting}[language=c]
<(X1 * Y1) --> (ocr * ocr)> <=> <(Y1 * X1) --> (ocr * ocr)>

<($1 * $2) --> (ocr * ocr)> <=> <($2 * $1) --> (ocr * ocr)>
\end{lstlisting}

\subsubsection{Learning Transitivity}

With experience of the $X1 \rightarrow Y1$, $Y1 \rightarrow Z1$ and $X1 \rightarrow Z1$ relations (in the matching-to-sample), the system will be trained explicitly on transitivity. The following will be derived:

\begin{lstlisting}[language=c]
<(X1 * Y1) --> (ocr * ocr)> && <(Y1 * Z1) --> (ocr * ocr)> ==> 
 <(X1 * Z1) --> (ocr * ocr)>

<($1 * $2) --> (ocr * ocr)> && <($2 * $3) --> (ocr * ocr)> ==> 
 <($1 * $3) --> (ocr * ocr)>
\end{lstlisting}

\subsubsection{Learning a Relation From Functional Equivalence} \label{paragraph:learn_func}

As described in Section 4 of the main paper, the idea of acquiring relations can be generalized to any procedure. Here it is assumed to be possible also in a functional equivalence procedure. 

First, two three-term contingencies are learned, where $X1$ and $Y1$ both function as a discriminative stimulus. 

\begin{lstlisting}[language=c]
<(sample * X1) --> (loc * ocr)>. :|:
<({SELF} * op1) --> ^action>. :|:
G. :|:

100 // Wait 100 time steps

<(sample * Y1) --> (loc * ocr)>. :|:
<({SELF} * op1) --> ^action>. :|:
G. :|:
\end{lstlisting}

A functional equivalence between the two $X1$ and $Y1$ events that had the same functional role (discriminative stimulus) would be obtained:

\begin{lstlisting}[language=c]
<(sample * X1) --> (loc * ocr)> <=> 
 <(sample * Y1) --> (loc * ocr)>
\end{lstlisting}

Importantly, the system could also acquire a relation of this kind:

\begin{lstlisting}[language=c]
<(X1 * Y1) --> (ocr * ocr)> && 
 <(sample * sample) --> (loc * loc)> 
 ==> 
 <(sample * X1) --> (loc * ocr)> <=> 
  <(sample * Y1) --> (loc * ocr)>
\end{lstlisting}

An abstract version of this would also be derived:

\begin{lstlisting}[language=c]
<($1 * $2) --> (ocr * ocr)> && <($3 * $3) --> (loc * loc)> ==> 
 <($3 * $1) --> (loc * ocr)> <=> 
 <($3 * $2) --> (loc * ocr)> 
\end{lstlisting}

\subsection{Phase 2: Training conditional discriminations}

The $ABC$ networks were trained using conditional discriminations with the matching-to-sample procedure. For example, the $A1 \rightarrow B1$ could be trained using feedback as follows:

\begin{lstlisting}[language=c]
<(sample * A1) --> (loc * ocr)>. :|:
<(left * B1) --> (loc * ocr)>. :|:
<(right * B2) --> (loc * ocr)>. :|:
// Executed with for example motor babbling
<({SELF} * (sample * left)) --> ^match>. :|:
G. :|:

<(sample * A1) --> (loc * ocr)> &/ <(left * B1) --> (loc * ocr)> 
 &/ <({SELF} * (sample * left)) --> ^match>) =/> G>

<(A1 * B1) --> (ocr * ocr)> && <(sample * left) --> (loc * loc)>
 ==>
<(sample * A1) --> (loc * ocr)> &/ <(left * B1) --> (loc * ocr)> 
 &/ <({SELF} * (sample * left)) --> ^match>) =/> G>
\end{lstlisting}

In all, the $A1-B1-C1$ network (Figure 4 in the main text) with the derivations would be as follows:

\begin{lstlisting}[language=c]
// (1) Trained
<(A1 * B1) --> (ocr * ocr)>

// (2) Trained
<(A1 * C1) --> (ocr * ocr)> 

// (3) Derived by symmetry from (1)
<(B1 * A1) --> (ocr * ocr)>
 
// (4) Derived by symmetry from (2)
<(C1 * A1) --> (ocr * ocr)>

// (5) Derived by transitivity from (3)+(2)
<(B1 * C1) --> (ocr * ocr)>

// (6) Derived by symmetry from (5)
<(C1 * B1) --> (ocr * ocr)>
\end{lstlisting}

The corresponding $A2-B2-C2$ network would also be derived.

\subsection{Phase 3: Function Training}

In this phase, two discriminative functions were trained for $B1$ and $B2$, respectively.

\begin{lstlisting}[language=c]
<(sample * B1) --> (loc * ocr)>. :|:

// Babbled or explicitly shown to the system
<({SELF} * clap) --> ^action>. :|:
G. :|:

// Derived:
<(sample * B1) --> (loc * ocr)>) &/ 
 <({SELF} * clap) --> ^action> =/> G>.

<(sample * B2) --> (loc * ocr)>. :|:
<({SELF} * wave) --> ^action>. :|:
G. :|:

// Derived:
<(sample * B2) --> (loc * ocr)>) &/ 
 <({SELF} * wave) --> ^action> =/> G>.
\end{lstlisting}

\subsection{Phase 4: Testing derived relations and transfer}

The first part of the testing, concerning matching-to-sample, would involve a situation like the following.

\begin{lstlisting}[language=c]
<(sample * C1) --> (loc * ocr)>. :|:
<(left * B1) --> (loc * ocr)>. :|:
<(right * B2) --> (loc * ocr)>. :|:
G! :|:
\end{lstlisting}

First, by interacting with the scene, the system is assumed to acquire the following relation (as described in Section 4 of the main text):

\begin{lstlisting}[language=c]
<(sample * left) --> (loc * loc)>
\end{lstlisting}

The following relation has been derived from before:

\begin{lstlisting}[language=c]
<(C1 * B1) --> (ocr * ocr)>
\end{lstlisting}

With the general form of conditionality from Section \ref{paragraph:learn_cond}, then the following could be derived:

\begin{lstlisting}[language=c]
<(C1 * B1) --> (ocr * ocr)> && <(sample * left) --> (loc * loc)>
 ==>
<(sample * C1) --> (loc * ocr)> &/ <(left * B1) --> (loc * ocr)>) &/ <({SELF} * (sample * left)) --> ^match>) =/> G>.
\end{lstlisting}

Hence, the operation \texttt{<(\{SELF\} * (sample * left)) -{}-> \^{}match>} will be executed.

In the second part of testing, the following situation was presented to the system:

\begin{lstlisting}[language=c]
<(sample * C1) --> (loc * ocr)>. :|:
G! :|:
\end{lstlisting}

By interacting with the scene, the system is assumed to be able to acquire the following relation:

\begin{lstlisting}[language=c]
<(sample * sample) --> (loc * loc)>
\end{lstlisting}

With the derived parts of the network (including $C1 \rightarrow B1$) and the abstract form learned in Section \ref{paragraph:learn_func}, then the following would be derived:

\begin{lstlisting}[language=c]
<(C1 * B1) --> (ocr * ocr)> && 
 <(sample * sample) --> (loc * loc)> ==> 
<(sample * C1) --> (loc * ocr)> <=> 
<(sample * B1) --> (loc * ocr)>
\end{lstlisting}


then the operation \texttt{<(\{SELF\} * clap) -{}-> \^{}action>} would be executed, by substituting 

\begin{lstlisting}[language=c]
<(sample * C1) --> (loc * ocr)>. :|:
\end{lstlisting}

for 

\begin{lstlisting}[language=c]
<(sample * B1) --> (loc * ocr)>. :|:
\end{lstlisting}

and using the previously trained contingency:

\begin{lstlisting}[language=c]
<(sample * B1) --> (loc * ocr)>) &/ <({SELF} * clap) --> ^action>) =/> G>
\end{lstlisting}

Similarly, the following situation

\begin{lstlisting}[language=c]
<(sample * C2) --> (loc * ocr)>. :|:
G! :|:
\end{lstlisting}

would lead to the operation \texttt{<(\{SELF\} * wave) -{}-> \^{}action>} being executed, using similar derivations.

%

\section{Opposition and Transformation of Function}

\subsection{Phase 1: Pretraining of Relational Frames}

\subsubsection{Learning the Acquired Relations} \label{paragraph:learn_acq2}

In the example below, \texttt{rel} is also used as a location (Extending Figure 1 of the main paper). Initial training of the relation form would be trained using the matching-to-sample procedure:

\begin{lstlisting}[language=c]
<(rel * SAME) --> (loc * ocr)>. :|:
<(sample * X1) --> (loc * ocr)>. :|:
<(left * Y1) --> (loc * ocr)>. :|:
<(right * Y2) --> (loc * ocr)>. :|:
G! :|:
\end{lstlisting}

When learning (with feedback) to match the sample and the left option, the following contingency would be formed:

\begin{lstlisting}[language=c]
<(rel * SAME) --> (loc * ocr)> &/ 
 <(sample * X1) --> (loc * ocr)> &/ 
 <(left * Y1) --> (loc * ocr)> &/ 
 <({SELF} * (sample * left)) --> ^match>) =/> G>
\end{lstlisting}

An acquired relation SAME $X1 \rightarrow Y1$ (and a $(rel \times (sample \times left))$ relation) with corresponding implication will be formed:

\begin{lstlisting}[language=c]
<(SAME * (X1 * Y1)) --> (ocr * (ocr * ocr))> &&
 <(rel * (sample * left)) --> (loc * (loc * loc))>
 ==>
 <(rel * SAME) --> (loc * ocr)> &/ 
  <(sample * X1) --> (loc * ocr)> &/ 
  <(left * Y1) --> (loc * ocr)> &/ 
  <({SELF} * (sample * left)) --> ^match>) =/> G>
\end{lstlisting}

The abstract version that is also derived, with variables introduced, would be the following:

\begin{lstlisting}[language=c]
<($1 * ($2 * $3)) --> (ocr * (ocr * ocr))> &&
 <($4 * ($5 * $6)) --> (loc * (loc * loc))>
 ==>
 <($4 * $1) --> (loc * ocr)> &/ 
  <($5 * $2) --> (loc * ocr)> &/ 
  <($6 * $3) --> (loc * ocr)> &/ 
  <({SELF} * ($5 * $6)) --> ^match>) =/> G>
\end{lstlisting}

\subsubsection{Learning Mutual Entailment of SAME and OPPOSITE} \label{paragraph:learn_opp_mut}

The system will be explicitly trained on the SAME $X1 \rightarrow Y1$ and SAME $Y1 \rightarrow X1$ relations, using the matching-to-sample.

After training, mutual entailment for the SAME relation would be formed. First:

\begin{lstlisting}[language=c]
<(SAME * (X1 * Y1)) --> (ocr * (ocr * ocr))> && 
 <(rel * (sample * left)) --> (loc * (loc * loc))>
 <=>
<(SAME * (Y1 * X1)) --> (ocr * (ocr * ocr))> && 
 <(rel * (sample * left)) --> (loc * (loc * loc))>
\end{lstlisting}

This would be further reduced to:

\begin{lstlisting}[language=c]
<(SAME * (X1 * Y1)) --> (ocr * (ocr * ocr))>
 <=>
<(SAME * (Y1 * X1)) --> (ocr * (ocr * ocr))> 
\end{lstlisting}

The abstract form would be:

\begin{lstlisting}[language=c]
<($1 * ($2 * $3)) --> (ocr * (ocr * ocr))>
 <=>
<($1 * ($3 * $2)) --> (ocr * (ocr * ocr))> 
\end{lstlisting}

For OPPOSITE $X1 \rightarrow Y2$ and OPPOSITE $Y2 \rightarrow X1$ the representation of mutual entailment would be the same. It is only at the level of combinatorial entailment (see next paragraph) where these patterns differ.

\subsubsection{Learning Combinatorial Entailment} \label{paragraph:learn_opp_comb}

The following four concrete and four abstract patterns of combinatorial entailment would be explicitly trained.

For SAME-SAME combinations:

\begin{lstlisting}[language=c]
<(SAME * (X1 * Y1)) --> (ocr * (ocr * ocr))> && 
 <(SAME * (Y1 * Z1)) --> (ocr * (ocr * ocr))> ==>
  <(SAME * (X1 * Z1)) --> (ocr * (ocr * ocr))>

<($1 * ($2 * $3)) --> (ocr * (ocr * ocr))> && 
 <($1 * ($3 * $4)) --> (ocr * (ocr * ocr))> ==>
  <($1 * ($2 * $4)) --> (ocr * (ocr * ocr))>
\end{lstlisting}

For SAME-OPPOSITE combinations:

\begin{lstlisting}[language=c]
<(SAME * (X1 * Y1)) --> (ocr * (ocr * ocr))> && 
 <(OPPOSITE * (Y1 * Z2)) --> (ocr * (ocr * ocr))> ==>
  <(OPPOSITE * (X1 * Z2)) --> (ocr * (ocr * ocr))>

<(SAME * ($1 * $2)) --> (ocr * (ocr * ocr))> && 
 <($3 * ($2 * $4)) --> (ocr * (ocr * ocr))> ==>
  <($3 * ($1 * $4)) --> (ocr * (ocr * ocr))>
\end{lstlisting}

For OPPOSITE-SAME combinations:

\begin{lstlisting}[language=c]
<(OPPOSITE * (X1 * Y2)) --> (ocr * (ocr * ocr))> && 
 <(SAME * (Y2 * Z2)) --> (ocr * (ocr * ocr))> ==>
  <(OPPOSITE * (X1 * Z2)) --> (ocr * (ocr * ocr))>
  
<($1 * ($2 * $3)) --> (ocr * (ocr * ocr))> && 
 <(SAME * ($3 * $4)) --> (ocr * (ocr * ocr))> ==>
  <($1 * ($2 * $4)) --> (ocr * (ocr * ocr))>
\end{lstlisting}

For OPPOSITE-OPPOSITE combinations:

\begin{lstlisting}[language=c]
<(OPPOSITE * (X1 * Y2)) --> (ocr * (ocr * ocr))> && 
 <(OPPOSITE * (Y2 * Z1)) --> (ocr * (ocr * ocr))> ==>
  <(SAME * (X1 * Z1)) --> (ocr * (ocr * ocr))>
  
<($1 * ($2 * $3)) --> (ocr * (ocr * ocr))> && 
 <($1 * ($3 * $4)) --> (ocr * (ocr * ocr))> ==>
  <(SAME * ($2 * $4)) --> (ocr * (ocr * ocr))>
\end{lstlisting}

\subsubsection{Learning a Relation From Functional Equivalence} \label{paragraph:learn_abstfunc1}

For the system to be able to generalize from the Matching-to-sample procedure to a network of discriminative functions, the following was presented as part of the pretraining.

\begin{lstlisting}[language=c]
<(rel * SAME) --> (loc * ocr)>. :|:
<(sample * X1) --> (loc * ocr)>. :|:
<({SELF} * op1) --> ^action>. :|:
G. :|:

100 // Wait 100 time steps

<(sample * Y1) --> (loc * ocr)>. :|:
<({SELF} * op1) --> ^action>. :|:
G. :|:
\end{lstlisting}

A functional equivalence between the preconditions that functioned as discriminative stimuli would be obtained:

\begin{lstlisting}[language=c]
<(rel * SAME) --> (loc * ocr)> &/ <(sample * X1) --> (loc * ocr)>
 <=>
<(sample * Y1) --> (loc * ocr)>
\end{lstlisting}

Second, an acquired relation would be derived, with the corresponding implication:

\begin{lstlisting}[language=c]
<(SAME * (X1 * Y1)) --> (ocr * (ocr * ocr))> && 
 <(rel * (sample * sample)) --> (loc * (loc * loc))>
 ==>
 <(rel * SAME) --> (loc * ocr)> &/ 
  <(sample * X1) --> (loc * ocr)>
  <=>
 <(sample * Y1) --> (loc * ocr)>
\end{lstlisting}

The abstract form would be the following:

\begin{lstlisting}[language=c]
<($1 * ($2 * $3)) --> (ocr * (ocr * ocr))> && 
 <($4 * ($5 * $5)) --> (loc * (loc * loc))>
 ==>
 <($4 * $1) --> (loc * ocr)> &/ <($5 * $2) --> (loc * ocr)>
  <=>
 <($5 * $3) --> (loc * ocr)>
\end{lstlisting}

Regarding learning the OPPOSITE relation with discriminative functions, the following was presented.

\begin{lstlisting}[language=c]
<(rel * OPPOSITE) --> (loc * ocr)>. :|:
<(sample * X1) --> (loc * ocr)>. :|:
<({SELF} * op2) --> ^action>. :|:
G. :|:

100 // Wait 100 time steps

<(sample * Y2) --> (loc * ocr)>. :|:
<({SELF} * op2) --> ^action>. :|:
G. :|:
\end{lstlisting}

A functional equivalence would be obtained:

\begin{lstlisting}[language=c]
<(rel * OPPOSITE) --> (loc * ocr)> &/ <(sample * X1) --> (loc * ocr)>
 <=>
<(sample * Y2) --> (loc * ocr)>
\end{lstlisting}

Then, an acquired relation would be derived, with a corresponding implication:

\begin{lstlisting}[language=c]
<(OPPOSITE * (X1 * Y2)) --> (ocr * (ocr * ocr))> && 
 <(rel * (sample * sample)) --> (loc * (loc * loc))>
 ==>
 <(rel * OPPOSITE) --> (loc * ocr)> &/ 
  <(sample * X1) --> (loc * ocr)>
  <=>
 <(sample * Y2) --> (loc * ocr)>
\end{lstlisting}

The abstract form would be:

\begin{lstlisting}[language=c]
<($1 * ($2 * $3)) --> (ocr * (ocr * ocr))> && 
 <($4 * ($5 * $5)) --> (loc * (loc * loc))>
 ==>
 <($4 * $1) --> (loc * ocr)> &/ 
  <($5 * $2) --> (loc * ocr)>
  <=>
 <($5 * $3) --> (loc * ocr)>
\end{lstlisting}

Importantly, this is the same abstract form as for that of the SAME relation.

\subsection{Phase 2: Training of Relational Networks}

In this phase, it will be illustrated how the network (as illustrated in Figure 5 in the main text) will be trained using the Matching-to-sample procedure, with SAME and OPPOSITE cues:

\begin{lstlisting}[language=c]
<(rel * SAME) --> (loc * ocr)>. :|:
<(sample * A1) --> (loc * ocr)>. :|:
<(left * B1) --> (loc * ocr)>. :|:
<(right * B2) --> (loc * ocr)>. :|:
G! :|:
\end{lstlisting}

After motor babbling, and feedback, the system would be trained into the following:

\begin{lstlisting}[language=c]
<(SAME * (A1 * B1)) --> (ocr * (ocr * ocr))> &&
 <(rel * (sample * left)) --> (loc * (loc * loc))>
 ==>
 <(rel * SAME) --> (loc * ocr)> &/ 
  <(sample * A1) --> (loc * ocr)> &/ 
  <(left * B1) --> (loc * ocr)> &/ 
  <({SELF} * (sample * left)) --> ^match>) =/> G>
\end{lstlisting}

After complete training of the entire relational network, the system would have been trained in, and having derived the following relations using the mutual and combinatorial entailment derived in Sections \ref{paragraph:learn_opp_mut} and \ref{paragraph:learn_opp_comb}:

\begin{lstlisting}[language=c]
// (1) Trained
<(SAME * (A1 * B1)) --> (ocr * (ocr * ocr))>

// (2) Trained
<(SAME * (A1 * C1)) --> (ocr * (ocr * ocr))>
 
// (3) Trained
<(OPPOSITE * (A1 * B2)) --> (ocr * (ocr * ocr))>

 // (4) Trained
<(OPPOSITE * (A1 * C2)) --> (ocr * (ocr * ocr))>

 // (5) Derived by mutual entailment from (1)
<(SAME * (B1 * A1)) --> (ocr * (ocr * ocr))>
 
// (6) Derived by mutual entailment from (2)
<(SAME * (C1 * A1)) --> (ocr * (ocr * ocr))>

// (7) Derived by combinatorial entailment from (5)+(2)
<(SAME * (B1 * C1)) --> (ocr * (ocr * ocr))>

// (8) Derived by mutual entailment from (7)
<(SAME * (C1 * B1)) --> (ocr * (ocr * ocr))>

// (9) Derived by combinatorial entailment from (6)+(3)
<(OPPOSITE * (C1 * B2)) --> (ocr * (ocr * ocr))>

// (10) Derived by mutual entailment from (3)
<(OPPOSITE * (B2 * A1)) --> (ocr * (ocr * ocr))>
 
// (11) Derived by mutual entailment from (4)
<(OPPOSITE * (C2 * A1)) --> (ocr * (ocr * ocr))>

// (12) Derived by combinatorial entailment from (11)+(1)
<(OPPOSITE * (C2 * B1)) --> (ocr * (ocr * ocr))>

// (13) Derived by combinatorial entailment from (10)+(4)
<(SAME * (B2 * C2)) --> (ocr * (ocr * ocr))>
 
// (14) Derived by mutual entailment from (13)
<(SAME * (C2 * B2)) --> (ocr * (ocr * ocr))>

\end{lstlisting}

\subsection{Phase 3: Function Training}

In this phase, two discriminative functions were trained for $B1$ and $B2$, respectively.

\begin{lstlisting}[language=c]
<(sample * B1) --> (loc * ocr)>. :|:
<({SELF} * clap) --> ^action>. :|:
G. :|:

// Derived:
<(sample * B1) --> (loc * ocr)> &/ 
 <({SELF} * clap) --> ^action> =/> G>.

<(sample * B2) --> (loc * ocr)>. :|:
<({SELF} * wave) --> ^action>. :|:
G. :|:

// Derived:
<(sample * B2) --> (loc * ocr)> &/ 
 <({SELF} * wave) --> ^action> =/> G>.
\end{lstlisting}

\subsection{Phase 4: Testing Derived Relations and Function Transformations}

The first part of the testing would involve a situation like the following, where for example the derived relations SAME $C1 \rightarrow B1$ and OPPOSITE $C1 \rightarrow B1$ would be tested.

\begin{lstlisting}[language=c]
<(rel * SAME) --> (loc * ocr)>. :|:
<(sample * C1) --> (loc * ocr)>. :|:
<(left * B1) --> (loc * ocr)>. :|:
<(right * B2) --> (loc * ocr)>. :|:
G! :|:
\end{lstlisting}

First, by interacting with the scene, it is assumed that the system would acquire the following relation:

\begin{lstlisting}[language=c]
<(rel * (sample * left)) --> (loc * (loc * loc))>
\end{lstlisting}

The following relation has been derived from before:

\begin{lstlisting}[language=c]
<(SAME * (C1 * B1)) --> (ocr * (ocr * ocr))>
\end{lstlisting}

With the abstract form from Section \ref{paragraph:learn_acq2}, then the following could be derived:

\begin{lstlisting}[language=c]
<(SAME * (C1 * B1)) --> (ocr * (ocr * ocr))> && 
 <(rel * (sample * left)) --> (loc * (loc * loc))>
 ==>
<(rel * SAME) --> (loc * ocr)> &/ <(sample * C1) --> (loc * ocr)> &/ 
 <(left * B1) --> (loc * ocr)>) &/ 
 <({SELF} * (sample * left)) --> ^match>) =/> G>.
\end{lstlisting}

Hence, the operation \texttt{<(\{SELF\} * (sample * left)) -{}-> \^{}match>\\} would be executed.

In the second part of the testing, the following situation was presented to the system (Bottom row of Figure 3 in the main text):

\begin{lstlisting}[language=c]
<(rel * SAME) --> (loc * ocr)>. :|:
<(sample * C1) --> (loc * ocr)>. :|:
G! :|:
\end{lstlisting}

Assuming that the system, by interacting with the situation (as described in Section 4 of the main text) would acquire:

\begin{lstlisting}[language=c]
<(rel * (sample * sample)) --> (loc * (loc * loc))>
\end{lstlisting}

The system would also use the following previously derived relation:

\begin{lstlisting}[language=c]
<(SAME * (C1 * B1)) --> (ocr * (ocr * ocr))>
\end{lstlisting}

Those two relations would derive the following functional equivalence, using the abstract form for SAME as described in Section \ref{paragraph:learn_abstfunc1}:

\begin{lstlisting}[language=c]
<(SAME * (C1 * B1)) --> (ocr * (ocr * ocr))> && 
 <(rel * (sample * sample)) --> (loc * (loc * loc))>
 ==>
 <(rel * SAME) --> (loc * ocr)> &/ <(sample * C1) --> (loc * ocr)>
  <=>
 <(sample * B1) --> (loc * ocr)>
\end{lstlisting}

Then, by substituting 

\begin{lstlisting}[language=c]
<(rel * SAME) --> (loc * ocr)> &/ 
 <(sample * C1) --> (loc * ocr)>
\end{lstlisting}

for 

\begin{lstlisting}[language=c]
<(sample * B1) --> (loc * ocr)>. :|:
\end{lstlisting}

the operation \texttt{<(\{SELF\} * clap) -{}-> \^{}action>} would be executed, by using the previously trained contingency involving $B1$ and \texttt{\^{}clap}.

Similarly, the following situation

\begin{lstlisting}[language=c]
<(rel * OPPOSITE) --> (loc * ocr)>. :|:
<(sample * C1) --> (loc * ocr)>. :|:
G! :|:
\end{lstlisting}

would lead to the operation \texttt{<(\{SELF\} * wave) -{}-> \^{}action>} would be executed, using similar derivations. Especially the following network will be important in the derivation:

\begin{lstlisting}[language=c]
<(OPPOSITE * (C1 * B2)) --> (ocr * (ocr * ocr))> && 
 <(rel * (sample * sample)) --> (loc * (loc * loc))>
 ==>
 <(rel * OPPOSITE) --> (loc * ocr)> &/ 
  <(sample * C1) --> (loc * ocr)>
  <=>
 <(sample * B2) --> (loc * ocr)>
\end{lstlisting}

\clearpage

\section*{Figure captions}

\begin{figure}[h!]
\centering
\includegraphics[scale=0.17]{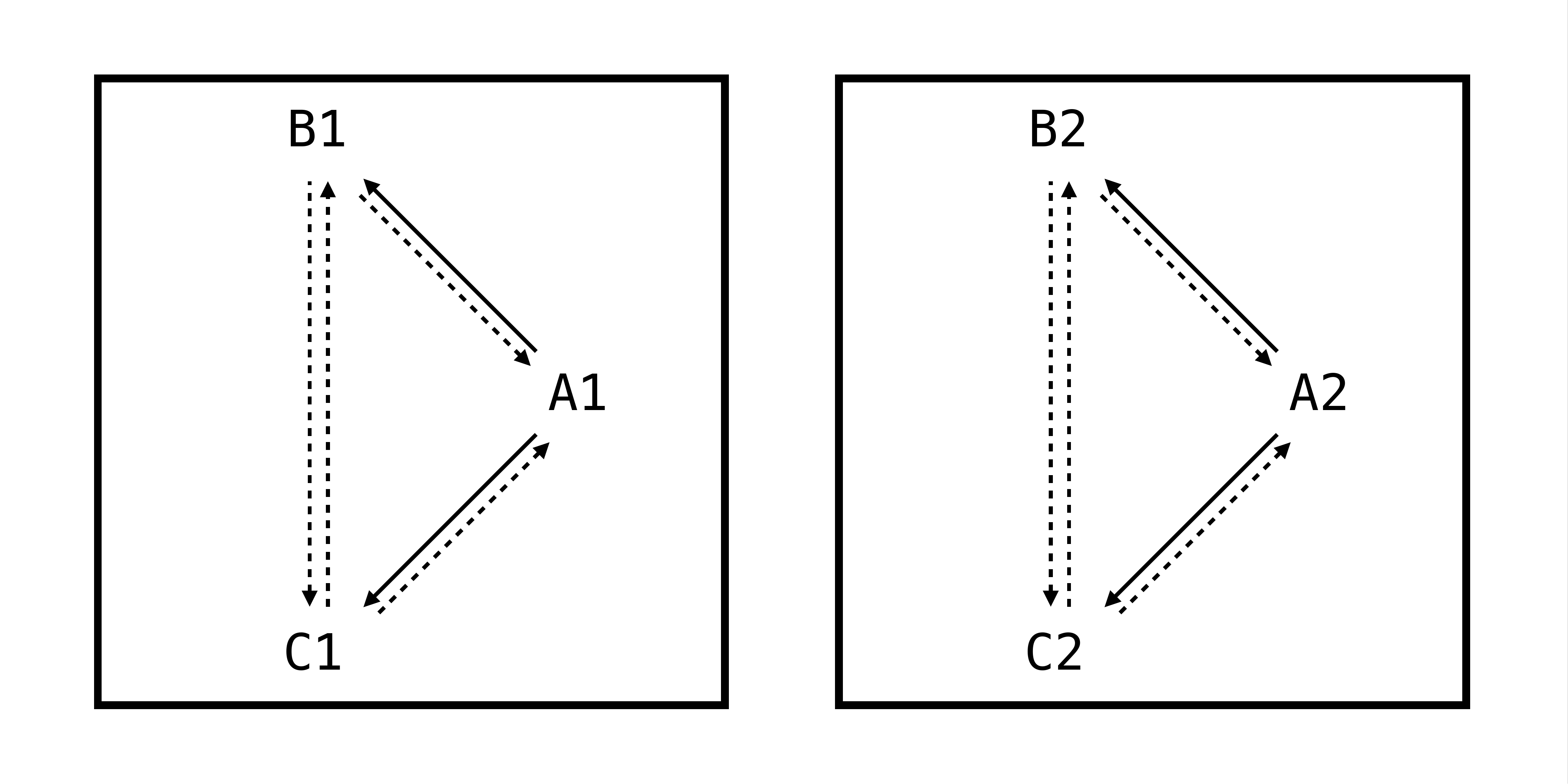}
\caption{
The two networks trained as part of the first experiment of this paper. Solid arrows represent relations that are explicitly trained. Dashed arrows represent derived relations.} 
\label{fig_abc_same}
\end{figure}

\begin{figure}[h!]
\centering
\includegraphics[scale=0.18]{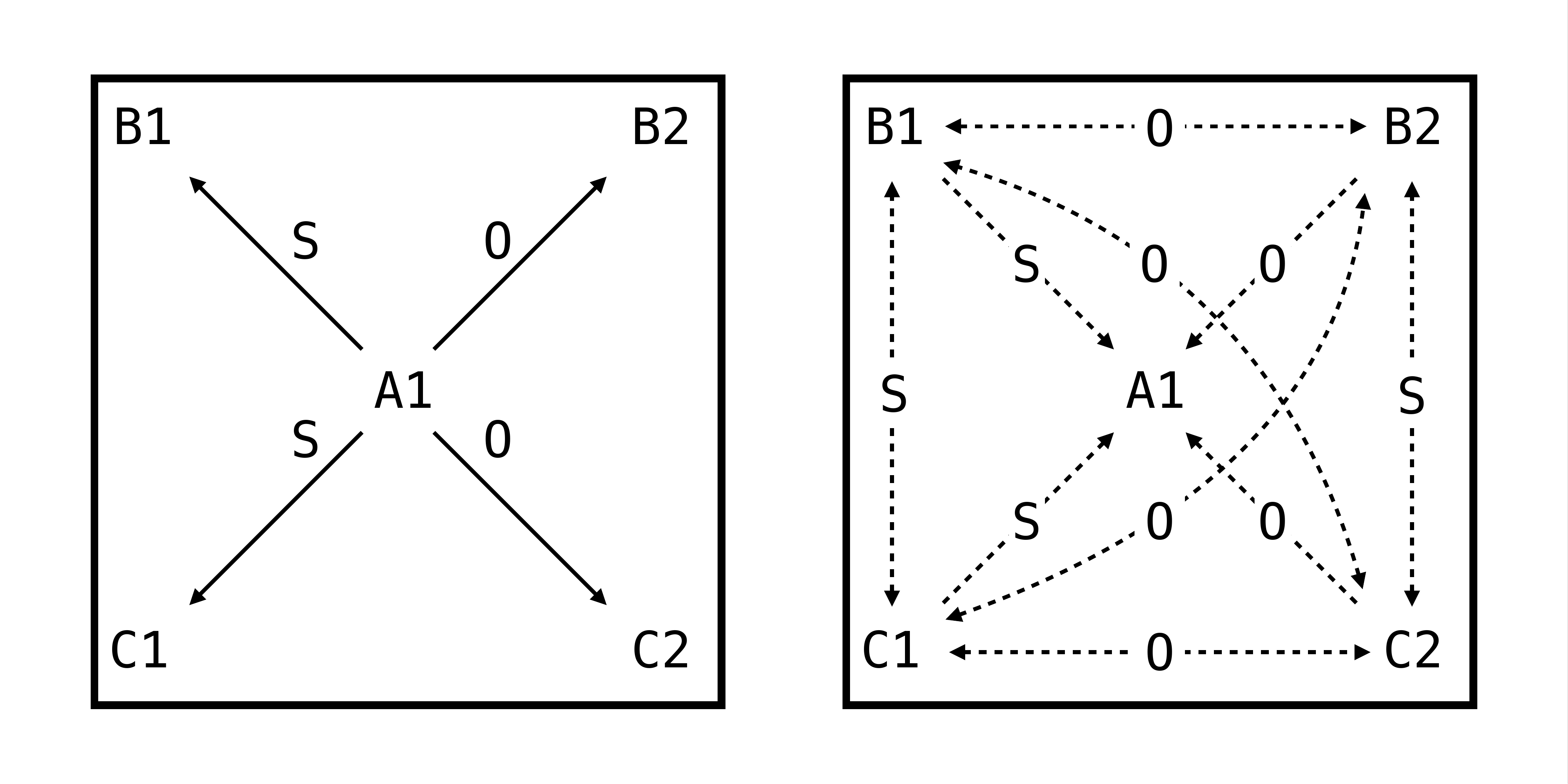}
\caption{
The network trained as part of the second experiment of this paper. S and O indicate SAME and OPPOSITE, respectively. Left panel shows relations that are explicitly trained. Right panel shows derived relations.} 
\label{fig_abc_opposite}
\end{figure}

\clearpage 

\section*{Tables}

\begin{table}[htbp]
    \centering
    \caption{Overview of Psychological Processes, NARS Mechanisms, Layers (from \cite{wang2013nalbook}), and References}
    \label{table_empirical_studies}
    \begin{tabular}{|l|l|c|l|}
        \hline
        \textbf{Psychological Process} & \textbf{NARS Mechanisms} & \textbf{NARS Layers} & \textbf{Reference} \\ \hline
        Operant Conditioning & \shortstack[l]{Temporal Reasoning and \\ Procedural Reasoning} & 7--8 & \citep{johansson2024machine} \\ \hline
        Generalized Identity Matching & +Abstraction & +6 & \citep{johansson2023generalized} \\ \hline
        Functional Equivalence & +Implications & +5 & \citep{johansson2024functional} \\ \hline
        \shortstack[l]{Arbitrarily Applicable \\ Relational Responding} & +Acquired Relations & +4 & This study \\ \hline
    \end{tabular}
    \label{table_empirical_studies}
\end{table}

\end{document}